%% file: sample-sigconf.tex
\documentclass[dvipsnames,sigconf=true, nonacm=false]{acmart}
\AtBeginDocument{%
  }

\setcopyright{acmlicensed}
\copyrightyear{2018}
\acmYear{2018}
\acmDOI{XXXXXXX.XXXXXXX}
\acmConference[Conference acronym 'XX]{Make sure to enter the correct
  conference title from your rights confirmation email}{June 03--05,
  2018}{Woodstock, NY}
\acmISBN{978-1-4503-XXXX-X/2018/06}

\usepackage{subcaption}
\usepackage{lipsum}
\usepackage{booktabs}
\usepackage{multirow}
\usepackage{framed}
\usepackage[most]{tcolorbox}
\usepackage{placeins}

\tcbset {
  base/.style={
    arc=0mm, 
    bottomtitle=0.5mm,
    boxrule=0mm,
    colbacktitle=black!10!white, 
    coltitle=black, 
    fonttitle=\bfseries, 
    left=2.5mm,
    leftrule=1mm,
    right=3.5mm,
    title={#1},
    toptitle=0.75mm, 
    breakable
  }
}

\definecolor{brandblue}{rgb}{0.34, 0.7, 1}
\newtcolorbox{mainbox}[1]{
  colframe=brandblue, 
  base={#1}
}

\newtcolorbox{subbox}[1]{
  colframe=black!30!white,
  base={#1}
}

\begin{document}

\title{Optimizing Photonic Structures with Large Language Model Driven Algorithm Discovery}


\author{Haoran Yin}
\email{h.yin@liacs.leidenuniv.nl}
\affiliation{%
  \institution{LIACS, Leiden University}
  \streetaddress{Einsteinweg 55}
  \city{Leiden}
  \country{Netherlands}
}

\author{Anna V. Kononova}
\email{a.kononova@liacs.leidenuniv.nl}
\orcid{0000-0002-4138-7024}
\affiliation{%
  \institution{LIACS, Leiden University}
  \streetaddress{Einsteinweg 55}
  \city{Leiden}
  \country{Netherlands}
}

\author{Thomas B{\"a}ck}
\email{t.h.w.baeck@liacs.leidenuniv.nl}
\orcid{0000-0001-6768-1478}
\affiliation{%
  \institution{LIACS, Leiden University}
  \streetaddress{Einsteinweg 55}
  \city{Leiden}
  \country{Netherlands}
}

\author{Niki van Stein}
\email{n.van.stein@liacs.leidenuniv.nl}
\orcid{0000-0002-0013-7969}
\affiliation{%
  \institution{LIACS, Leiden University}
  \streetaddress{Einsteinweg 55}
  \city{Leiden}
  \country{Netherlands}
}

\renewcommand{\shortauthors}{Yin et al.}

\begin{abstract}
We study how large language models can be used in combination with evolutionary computation techniques to automatically discover optimization algorithms for the design of photonic structures. Building on the Large Language Model Evolutionary Algorithm (LLaMEA) framework, we introduce structured prompt engineering tailored to multilayer photonic problems such as Bragg mirror, ellipsometry inverse analysis, and solar cell antireflection coatings. We systematically explore multiple evolutionary strategies, including (1+1), (1+5), (2+10), and others, to balance exploration and exploitation. Our experiments show that LLM-generated algorithms, generated using small-scale problem instances, can match or surpass established methods like quasi-oppositional differential evolution on large-scale realistic real-world problem instances. Notably, LLaMEA's self-debugging mutation loop, augmented by automatically extracted problem-specific insights, achieves strong anytime performance and reliable convergence across diverse problem scales. This work demonstrates the feasibility of domain-focused LLM prompts and evolutionary approaches in solving optical design tasks, paving the way for rapid, automated photonic inverse design.
\end{abstract}




\maketitle

\section{Introduction}
\label{sec:Introduction}
Optimization of photonic structures plays a key role in the advancement of technologies in various fields such as telecommunication, solar energy, and materials science~\cite{zhan2023light,abdellatif2019optimizing,liu2015photonic,long2009design,kim2024datasets}. However, the complexity and high dimensionality of these optimization problems pose significant challenges. Recent developments in automatic algorithm design, especially those that utilize large language models (LLMs), offer promising solutions.

LLMs have become powerful tools in the field of algorithm discovery and optimization. Several studies have demonstrated the ability of LLMs to automatically generate and refine optimization algorithms through an iterative process~\cite{liu2024evolution,van2024llamea}. By combining the power of large-scale language models with automatic algorithm design, these approaches open new avenues for developing algorithms for complex problems without relevant expertise.

In this work, we extend the capabilities of the Large Language Model Evolutionary Algorithm (LLaMEA) framework~\cite{van2024llamea} to deal with real-world photonic problems by addressing two critical limitations:
\begin{itemize}
    \item \textbf{Generic Task Prompts}: Original LLaMEA prompts lacked domain-specific guidance, which can lead to suboptimal algorithm designs for photonic problems.
    \item \textbf{Limited Evolutionary Strategy Diversity}: Previous studies focused only on simple (1,1) and (1+1) strategies, neglecting the potential of population-based exploration.
\end{itemize}
To overcome these limitations, we introduce structured task prompts enriched with photonics-specific problem descriptions and algorithmic insights. Furthermore, we systematically evaluate five new evolutionary strategy configurations-(1,5), (1+5), (2,10), (2+10), and (5+5)-to balance exploration-exploitation trade-offs.

The paper is organized as follows. Sec.~\ref{sec:Related Work} reviews the related work on LLM-driven algorithm discovery and photonic structure optimization problems. Sec.~\ref{sec:methodology} describes the methodology in detail. Sec.~\ref{sec:setup} provides data related to the experimental setup. Sec.~\ref{sec:results} shows the experimental results. Sec.~\ref{sec:Conclusions} summarizes the results of the work and presents future work.

\section{Related Work}
\label{sec:Related Work}
The rapid development of LLMs and optimization techniques has stimulated a strong interest in combining LLM-driven algorithm discovery with domain-specific applications. This section provides an overview of related work in two key areas: algorithmic automatic generation tools using LLMs and photonic structure optimization problems.

\subsection{Large Language Models for Algorithm Discovery}
Recent research has explored how LLMs can contribute to algorithmic discovery~\cite{liu2024systematic}. Two state-of-the-art representative frameworks in this area are Evolution of Heuristic (EoH) and LLaMEA~\cite{liu2024evolution,van2024llamea}. Both frameworks illustrate the potential of LLMs to automate and accelerate algorithmic discovery.

EoH combines LLMs with evolutionary computation to iteratively generate and refine heuristics. These heuristics are small functions that are then inserted into a larger code template before the evaluation of a problem. It evolves both natural-language descriptions of heuristics and executable code representations of these heuristics, allowing for diverse exploration and efficient improvement. EoH demonstrates state-of-the-art performance in heuristic automated design on combinatorial optimization problems, outperforming existing methods such as FunSearch on bin-packing and traveling salesperson problems~\cite{liu2024evolution}.

LLaMEA also uses LLMs within an evolution algorithm framework, focusing mainly on the (1,1) and (1+1) evolutionary strategies, to automatically generate, mutate and optimize complete metaheuristics~\cite{van2024llamea,rechenberg1978evolutionsstrategien}. Unlike EoH's emphasis on `thought', LLaMEA directly optimizes the algorithm itself, using run-time performance as feedback for iterative improvement. In benchmarking continuous problems, it produces algorithms that are comparable to or better than state-of-the-art optimization methods such as the covariance matrix adaptation evolution strategy (CMA-ES) and differential evolution (DE)~\cite{van2024llamea}.
Furthermore, the recently proposed LLaMEA-HPO framework augments LLaMEA with automated hyperparameter tuning, offloading parameter optimization from the LLM and thereby boosting algorithm quality while reducing the overall LLM query cost~\cite{van2024loop}.

While both EoH and LLaMEA are open-source and easy to extend, we chose to base this research on LLaMEA due to it's superior performance for discovering algorithms in the black-box optimization domain and for its ability to generate and optimize large code-bases.

\subsection{Earlier Methods for Automated Algorithm Discovery}

Beyond EoH and LLaMEA, a variety of other frameworks explore how LLMs can automate algorithm discovery. One of the earliest works is \textit{FunSearch}~\cite{FunSearch2024}, which applies an LLM within a distributed evolutionary search over function spaces. Starting from seed functions, FunSearch evolves them by prompting the LLM for refined or alternative code. 

(AEL)~\cite{liu2023algorithm} and \textit{ReEvo}~\cite{ye2024reevo}, similarly integrate LLMs within an iterative loop, generating small heuristic code snippets and improving them via mutation and crossover prompts. While EoH focuses on heuristics in text and code form, AEL/ReEvo explore a closely related direction by evolving coded modules. Both frameworks have shown promising results on combinatorial benchmarks such as the Traveling Salesperson Problem. 

\subsection{Optimization of Photonic Structures}
Optimization of photonic structures has been an important area of research, as the need for high efficiency and high precision drives the optimization of photonic structures in various applications such as communication, semiconductors, LED displays, materials analysis and solar cells~\cite{zhan2023light,abdellatif2019optimizing,liu2015photonic,long2009design,kim2024datasets}.
The following are three real-world problems related to global optimization of multilayered photonic structures, all of which have vital applications.

\subsubsection*{Bragg Mirror}
A Bragg reflector mirror, or Bragg mirror, is a series of two or more semiconductors or dielectric materials stacked in a staggered pattern to achieve high reflectivity in a particular optical band~\cite{bragg1914mr}. It has a wide range of applications, such as the construction of acoustic wave reflectors and filters~\cite{priyadarshini2024distributed}, the analysis of the crystal structure of materials~\cite{mihai2015metallic}, the improvement of solar cell efficiency~\cite{jiang2018design}, the detection of changes in physical quantities such as temperature and pressure~\cite{gryga2022distributed}, and the enhancement of the interaction between light and matter in quantum computation and quantum information~\cite{malak2014beyond}.

\subsubsection*{Ellipsometry Inverse Problem}
Ellipsometry is a nondestructive optical measurement technique that can be used to measure the thickness of the thin film, the refractive index and the absorption coefficients~\cite{rothen1945ellipsometer}. In chip fabrication, the properties of thin films have a critical impact on circuit performance~\cite{zollner2013spectroscopic}. The solution of the inverse problem of ellipsometry can help to improve the accuracy of the production process~\cite{toomey2024tackling}. It can also be used in the new energy and chemical industry to study the properties of solar cell materials, nanostructured thin films, and chemical coatings~\cite{fujiwara2018spectroscopic}. Ellipsometry inverse problem solving can not only improve the data analysis method in materials science, but also improve the applicability of thin film technology in industrial production~\cite{gonccalves2002fundamentals,schubert2004infrared,hinrichs2018ellipsometry}.

\subsubsection*{Photovoltaic Problem}
This refers to the design of a sophisticated antireflection coating for solar cells - a multilayer thin film structure optimized to minimize optical reflections and maximize the absorption of the active layer of the cell. Such coatings typically consist of alternating dielectric layers with different refractive indices. By suppressing surface reflection losses, a well-designed antireflective coating can significantly improve the power conversion efficiency of a solar cell (bare silicon surfaces, for instance, reflect ~30\% of incident sunlight)~\cite{ji2022recent}. The difficulty lies in achieving broad-band, all-encompassing antireflection: single-layer coatings can only eliminate reflections in narrow bands, so multilayer or graded index designs are needed to reduce reflectivity across the entire broad solar spectrum. Optimizing such coatings is a complex photonic inverse design problem, with many local minima in performance due to wave interference effects at multiple wavelengths~\cite{bennet2024illustrated}.

\section{Optimizing Photonic Structures with LLaMEA}
\label{sec:methodology}
We apply LLaMEA to real-world problems published by P. Bennet et al. to automatically discover and implement photonic structure optimization algorithms~\cite{van2024llamea,bennet:hal-02613161}.

\subsection{Real-world Problem Simulation}
Python-based Multilayer Optics Optimization and Simulation Hub (PyMoosh) is a numerical toolkit for calculating the optical properties of multilayer structures~\cite{langevin2024pymoosh}. PyMoosh is especially suited for complex problems involving multilayer optical structures (e.g. structural analysis of metallic or metalized layers) and can be extended to support advanced optimization and inverse problem design.

Building on PyMoosh, P. Bennet et al. have developed a testbed for defining and solving photonic optimization problems~\cite{bennet:hal-02613161}.
Based on their work, we migrate the following problems to the IOHexperimenter platform~\cite{IOHexperimenter}.
\begin{enumerate}
    \item Optimization of a Bragg mirror.
    \item Solving of an ellipsometry inverse problem.
    \item Design of a sophisticated anti-reflection coating to optimize solar absorption in a photovoltaic solar cell.
\end{enumerate}
These three problems are all optimization problems for photonic structures and are the focus of the completed benchmark work~\cite{bennet2024illustrated}. Adequate benchmark data on commonly used optimization algorithms in this area are provided for reference. The landscapes of these problems in 2D are shown in Fig.~\ref{fig:landscape}. This figure is intended to visualize the problem only, and since the dimensionality of the problem is directly related to the number of layers of the photonic structure, the 2D form of some problems is not of any practical interest.

\input{input/landscape}

\subsection{Enhanced Task Prompt Design}
To improve LLaMEA's ability to address domain-specific problems, we have added two key sections to the original task prompt:
\begin{itemize}
    \item Problem Description: A concise summary of the photonic structure optimization task, including key parameters (e.g., range of layer thicknesses, and dielectric constant constraints) and physical objectives (e.g., reflectivity maximization).
    \item Algorithmic Insight: Domain knowledge guidance, such as 'Encourage algorithms to detect and preserve modular structures' or 'Encourage periodicity in solutions through customized cost functions or constraints'.
\end{itemize}
These structured suggestions could lead to more specialized algorithms that perform better for photon-specific requirements.

\subsection{Automatic Generation of Algorithms}
We chose LLaMEA as the algorithm discovery method in our experiments for the following reasons:
\begin{enumerate}
    \item Targeting continuous optimization problems: Our problems are continuous optimization problems, and LLaMEA provides a solution specifically for such problems. It uses automated generation and iterative optimization based on metaheuristic algorithms suitable for dealing with the optimization of continuous variables.
    \item Reliable performance: LLaMEA has been shown to perform better than EoH in several complex optimization tasks~\cite{van2024llamea}.
\end{enumerate}

In our work, we perform rapid validation and benchmarking of algorithms generated on small-scale problem instances, and apply them on more complex problem instances. Feedback such as convergence speed and anytime performance metrics are used as feedback for LLaMEA to iteratively improve algorithms. After many improvements, LLaMEA will provide a number of usable algorithms. The algorithms validated as the best will participate in benchmarking and comparison with other commonly used state-of-the-art optimization algorithms.

\input{input/instances}

\section{Experimental Setup}
\label{sec:setup}
The experiments are divided into two main parts: \textbf{discovery} and \textbf{benchmarking}. That is, LLaMEA is used to search for optimization algorithms, and the algorithms discovered with the best performance are benchmarked against other commonly used optimization algorithms in different instances of the problem. The following content details the experimental setup in terms of problem setup, performance metric, prompt setup for LLM, and benchmarking.

\subsection{Problem Setup}
\subsubsection{Bragg Mirror}
For the Bragg mirror optimization problem, we set the optimization objective to find a photonic structure that has the strongest reflectance for light with a wavelength of $600$ nm. The permittivity of the first material is $1.96$, the permittivity of the other material is $3.24$, and the maximum thickness of each layer is $218$ nm. The problem instance with $10$ layers of alternating materials will collectively be referred to as \textit{mini-Bragg}, and those with $20$ layers of alternating materials will collectively be referred to as \textit{Bragg}.
\subsubsection{Ellipsometry Inverse Problem}
Regarding the ellipsoidal inverse problem, the substrate material is gold. The parameters of the reflective material to be solved include thickness and permittivity. The minimum thickness is $50$ nm, the maximum thickness is $150$ nm, the minimum permittivity is $1.1$, and the maximum permittivity is $3.0$. The number of layers of material is $1$. The problem instance in the latter collectively is referred to as \textit{ellipsometry}.
\subsubsection{Photovoltaic Problem}
The optimization objective for the instances of the photovoltaic problem is to optimize the photonic structure to improve solar absorption, with the thickness of each material layer ranging from $30$ to $250$ nm. The permittivity is $2.0$ for the first material and $3.0$ for the other. The substrate material parameters are set as default. Similarly, we call an instance with 10 layers \textit{photovoltaic}, 20 layers \textit{big-photovoltaic}, and 32 layers \textit{huge-photovoltaic}.

All problem instances are set as \textbf{minimization} problems. The algorithm discovery processes are based only on smaller instances, and the subsequent benchmarks on the best algorithms are based on all instances. Table~\ref{tab:instances} shows the evaluation budget and summarizes the parameter settings described above.

\subsection{Performance Metric}
We use the metric suggested by LLaMEA, AOCC, as a feedback to the LLM. Eq.~\ref{eq:aocc} is the definition of AOCC:
\begin{equation}
\resizebox{0.9\linewidth}{!}{
    $AOCC(y_{a,f}) = \frac{1}{B}\sum_{i=1}^{B} \left ( 1 - \frac{\min(\max((y_i),lb),ub) - lb}{ub-lb} \right )$
}
\label{eq:aocc}
\end{equation}
where $y_{a,f}$ is a series of log-scaled current-best fitness value of $f$ during the running of optimization algorithm $a$, $B$ is the evaluation budget, $y_i$ is the $i$-th element of $y_{a,f}$, $ub$ is the upper bound of $f$, and $lb$ is the lower bound of $f$.
In addition to the AOCC used by LLaMEA, we also add the optimal fitness value found at the end of the algorithm runs, $y^*$, to the feedback.

The algorithm discoveries are built on \textit{mini-Bragg}, \textit{ellipsometry}, and \textit{photovoltaic}, respectively. $ub$ is set to $1.0$ for \textit{mini-Bragg} and \textit{photovoltaic}, and $40.0$ for \textit{ellipsometry}.

\subsection{Prompt Setup}
The high degree of freedom in prompt customization is also a feature of LLaMEA, which uses the task prompt, the mutation prompt, the feedback prompt, and the output prompt, the first three of which are modified in our experiments.

Firstly, we state in the task prompt that the goal of the task is to find algorithms suitable for the optimization of multilayer photonic structures:
\begin{mainbox}{Task Prompt} 
    The optimization algorithm should be able to find high-performance solutions to a wide range of tasks, which include evaluation on real-world applications such as, e.g., optimization of multilayered photonic structures. \\
    \textit{<problem description>}\\
    \textit{<algorithmic insight>}\\
    Your task is to write the optimization algorithm in Python code. The code should contain an `\_\_init\_\_(self, budget, dim)` function and the function `def \_\_call\_\_(self, func)`, which should optimize the black box function `func` using `self.budget` function evaluations. 
    
    The func() can only be called as many times as the budget allows, not more. Each of the optimization functions has a search space between func.bounds.lb (lower bound) and func.bounds.ub (upper bound). The dimensionality can be varied.
    
    Give an excellent and novel heuristic algorithm to solve this task and also give it a one-line description with the main idea.
\end{mainbox}
where \textbf{problem description} and \textbf{algorithmic insight} are obtained by sending the following meta-prompts to gpt-4o-2024-08-06:
\begin{mainbox}{Meta-prompt for generating problem description}
    Please read \textit{Illustrated tutorial on global optimization in nanophotonics} first~\cite{bennet2024illustrated}. Then, give me summaries of bragg mirror problem, ellipsometry problem, and photovoltaics problem, respectively. The generated summaries need to be used in other chatgpt chats, so make sure they are understood by chatgpt.
\end{mainbox}
\begin{mainbox}{Meta-prompt for generating algorithmic insight}
    Please give me algorithmic insights of these three problems respectively. Similarly, these insights are used in other chatgpt chats, prompting chatgpt to generate optimization algorithms applicable to these problems, so make sure that these insights are understood by chatgpt.
\end{mainbox}
All descriptions and insights can be found in the supplementary material and the problem description and algorithmic insights for the Bragg mirror are as follows:
\begin{subbox}{Problem description for Bragg mirror}
\textbf{The Bragg mirror optimization} aims to maximize reflectivity at a wavelength of 600 nm using a multilayer structure with alternating refractive indices (1.4 and 1.8). The structure's thicknesses are varied to find the configuration with the highest reflectivity. The problem involves two cases: one with 10 layers (minibragg) and another with 20 layers (bragg), with the latter representing a more complex inverse design problem. The known optimal solution is a periodic Bragg mirror, which achieves the best reflectivity by leveraging constructive interference. This case exemplifies challenges such as multiple local minima in the optimization landscape. 
\end{subbox}
\begin{subbox}{Algorithmic insight for Bragg mirror}
    For this problem, the optimization landscape contains multiple local minima due to the wave nature of the problem. And periodic solutions are known to provide near-optimal results, suggesting the importance of leveraging constructive interference principles. Here are some suggestions for designing algorithms: 1. Use global optimization algorithms like Differential Evolution (DE) or Genetic Algorithms (GA) to explore the parameter space broadly. 2. Symmetric initialization strategies (e.g., Quasi-Oppositional DE) can improve exploration by evenly sampling the search space. 3. Algorithms should preserve modular characteristics in solutions, as multilayer designs often benefit from distinct functional blocks. 4. Combine global methods with local optimization (e.g., BFGS) to fine-tune solutions near promising regions. 5. Encourage periodicity in solutions via tailored cost functions or constraints. 
\end{subbox}
For each problem, we experiment with the task prompt without problem description and algorithmic insight, with problem description, and with both problem description and algorithmic insight. For each task prompt setting, LLaMEA runs 5 times with an evolutionary strategy (1 + 1), generating 100 algorithms per run. Each algorithm is executed $3$ times to eliminate experimental randomness.

Secondly, the dynamic mutation controlling prompt is set to control the LLaMEA mutation process, as this prompt has been validated to have better mutation control when combined with gpt-4o~\cite{yin2024controlling} with the fast mutation operator~\cite{doerr2017fast}:

\begin{mainbox}{Dynamic Mutation Controlling Prompt}
    Refine the strategy of the selected solution to improve it. Make sure that you only change $x\%$ of the code, which means if the code has 100 lines, you can only change $\lfloor x \rfloor$ lines, and the rest lines should remain the same. For this code, it has $n$ lines, so you can only change $\min(\lfloor n \times x / 100 \rfloor, 1)$ lines, the rest $n - \min(\lfloor n \times x / 100 \rfloor, 1)$ lines should remain the same. This changing rate $x\%$ is the mandatory requirement, you cannot change more or less than this rate.
\end{mainbox}
\noindent The mutation rate $x$ in the dynamic mutation control prompt is sampled from the heavy-tailed distribution known as the fast mutation operator~\cite{doerr2017fast}.

Finally, in the feedback prompt, we provide both AOCC and $y^*$ information to help LLM better establish the connection between algorithmic implementation and real-world results as a way to improve existing algorithms in a targeted manner:
\begin{mainbox}{Feedback Prompt}
    The algorithm \textless name\textgreater~got an average Area over the convergence curve (AOCC, 1.0 is the best) score of \textless aocc\_score\textgreater~with standard deviation \textless aocc\_std\textgreater. And the mean value of best solutions found was \textless y\_best\textgreater~(0. is the best) with standard deviation \textless y\_best\_std\textgreater.
\end{mainbox}

\subsection{Evolutionary Strategy Exploration}
After confirming which task prompt setting works best, we use that task prompt setting in the following experiments.

In addition to the initial (1,1) and (1+1) strategies, which are the default for LLaMEA, we investigate five other configurations of evolutionary strategies:
\begin{itemize}
    \item Comma strategy: (1,5), (2,10)
    \item The plus strategy: (1+5), (2+10), (5+5)
\end{itemize}
These configurations are often used in the black-box optimization literature, as they can balance exploration (by expanding the population of offspring) and exploitation (by retaining the elite), with $\lambda$ denoting the number of offspring and $\mu$ denoting the number of parents, and are indicated by $(\mu,\lambda)$~\cite{rechenberg1978evolutionsstrategien}.

For each of the configurations, LLaMEA runs $5$ times for each problem instance, and during each run, LLaMEA generates 100 algorithms. Each algorithm is executed $3$ times to eliminate experimental randomness.

\begin{figure}[!tb]
    \centering
    \begin{subfigure}[b]{0.9\linewidth}
        \includegraphics[width=\linewidth]{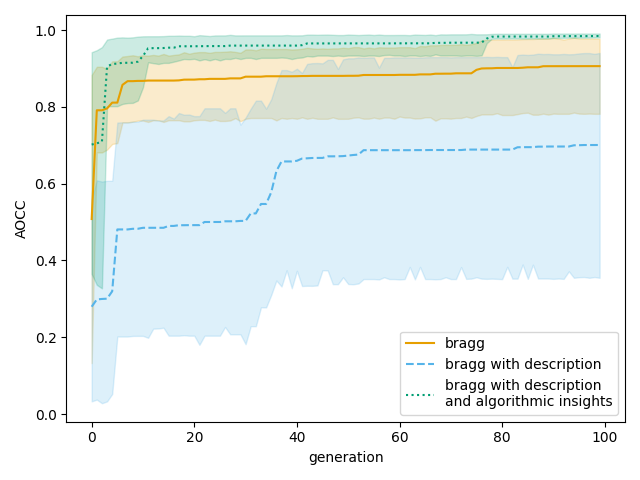}
        \caption{\textit{mini-Bragg} $\spadesuit$}
        \label{fig:auc_description_insight_bragg}
    \end{subfigure}
    \begin{subfigure}[b]{0.9\linewidth}
        \includegraphics[width=\linewidth]{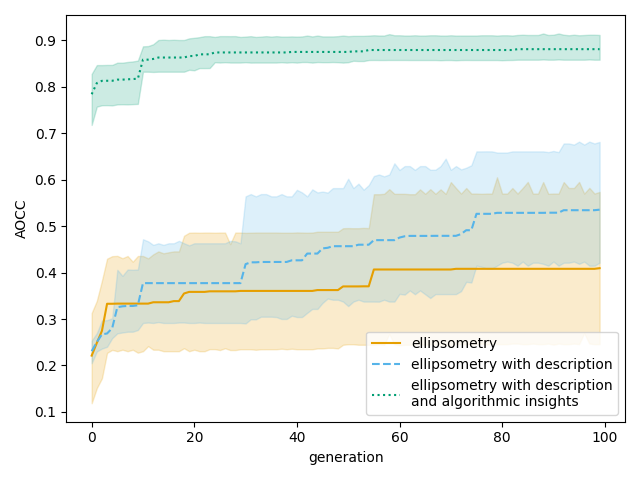}
        \caption{\textit{ellipsometry} $\spadesuit$}
        \label{fig:auc_description_insight_ellipsometry}
    \end{subfigure}
    \caption{Examples of task prompts that add problem descriptions and algorithm insights that improve the performance of LLaMEA. The higher the AOCC, the better. Prompts with descriptions and insights both outperform other prompt settings for \textit{mini-Bragg} and \textit{ellipsometry} instances.}
    \label{fig:description_insight_examples1}
\end{figure}

\subsection{Benchmarking}

After the algorithm discovery step, for each problem, we have $2500$ generated algorithms. Of these, the first 3 algorithms with the best AOCC average participate in the final benchmark. The optimal algorithms discovered through \textit{mini-Bragg} and \textit{photovoltaic} are also applied to higher-dimensional problem instances.
In addition, five algorithms commonly used in the field of photonic structure optimization are added to the benchmark to provide a performance comparison, including DE~\cite{feoktistov2006differential}, CMA-ES~\cite{hansen2003reducing}, Broyden–Fletcher–Goldfarb–Shanno (BFGS)~\cite{head1985broyden}, quasi-Newton differential evolution (QNDE)~\cite{noman2008accelerating}, and quasi-oppositional differential evolution (QODE)~\cite{rahnamayan2007quasi}.
For each problem instance, each algorithm is executed $15$ times.

\section{Results}
\label{sec:results}
This chapter presents the results of the experiments, for which the raw data and associated code are publicly available~\footnote{\url{https://doi.org/10.5281/zenodo.15073784}}.

\subsection{Enhanced Task Prompt Design}

\begin{figure}[!tb]
    \centering
    \begin{subfigure}[b]{0.9\linewidth}
        \includegraphics[width=\linewidth]{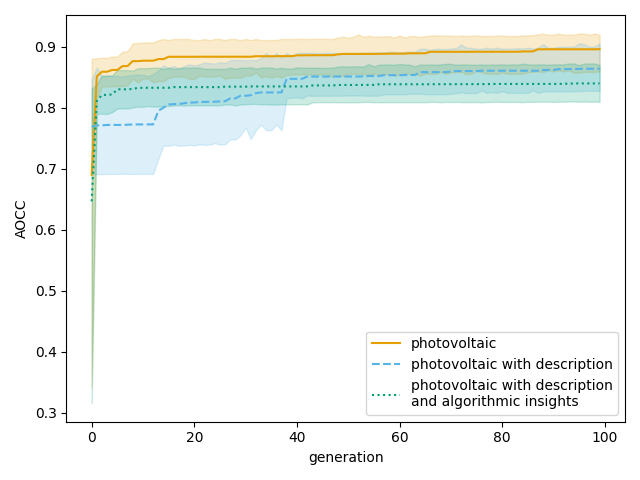}
        \caption{\textit{photovoltaic} $\spadesuit$}
        \label{fig:auc_description_insight_photovoltaic}
    \end{subfigure}
    \begin{subfigure}[b]{0.9\linewidth}
        \includegraphics[width=\linewidth]{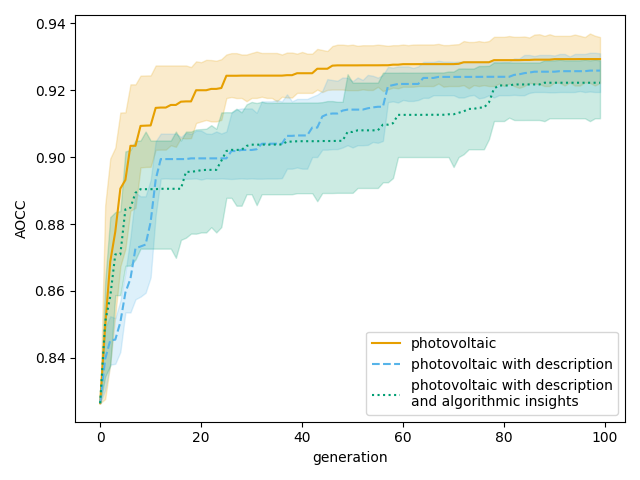}
        \caption{\textit{photovoltaic} $\spadesuit$, all runs start with the same solution.}
        \label{fig:auc_description_insight_photovoltaic_start_with_same_solution}
    \end{subfigure}
    \caption{Example of a task prompt that adds a problem description and algorithm insight that does not improve the effectiveness of LLaMEA. The higher the AOCC, the better. For \textit{photovoltaic} instance, task prompts without description and insight works best. Considering that the initial AOCC is high for all task prompt settings, Fig.~\ref{fig:auc_description_insight_photovoltaic_start_with_same_solution} shows the results for each LLaMEA run starting from the same solution to avoid the impact of initial gaps.}
    \label{fig:description_insight_examples2}
\end{figure}

The integration of domain-specific knowledge into structured prompts greatly influences the quality of the algorithms generated by LLaMEA. For the Bragg mirror designing and ellipsometry inverse problems, the detailed problem descriptions and algorithmic insights (see Appendix A) guided LLM to prioritize strategies that were consistent with physical principles. For example, prompts emphasizing 'periodic solutions' and 'constructive interference' motivate the algorithms to use modular initialization and symmetry constraints, which are essential for Bragg reflector optimization, as shown in Fig.~\ref{fig:auc_description_insight_bragg}. These prompts explicitly encourage the use of hybrid global-local search methods, e.g. combining DE and BFGS for local refinement, which improves AOCC by 10\% compared to general prompts. For ellipsometry, problem description and algorithmic insight provide more significant improvements.

\begin{figure*}[!tb]
    \centering
    \begin{subfigure}[b]{0.32\linewidth}
        \includegraphics[width=\linewidth, trim=0mm 0mm 0mm 0mm,clip]{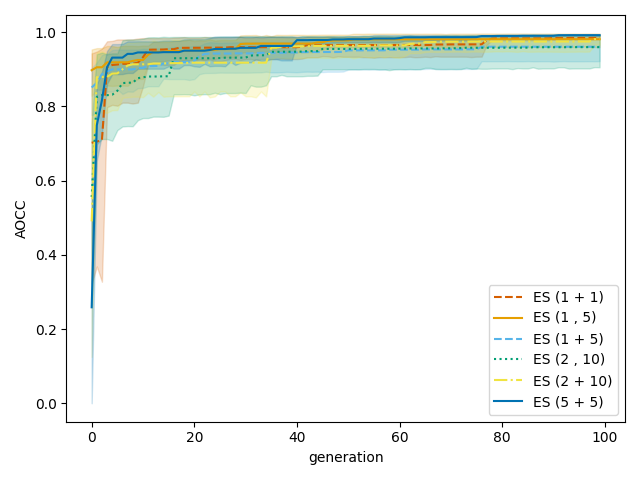}
        \caption{\textit{mini-Bragg} $\spadesuit$
        }
        \label{fig:auc_description_insight_bragg_population}
    \end{subfigure}
    \begin{subfigure}[b]{0.32\linewidth}
        \includegraphics[width=\linewidth, trim=0mm 0mm 0mm 0mm,clip]{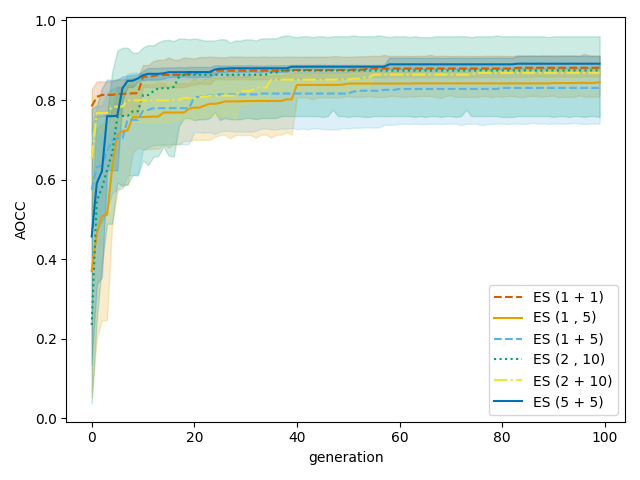}
        \caption{\textit{ellipsometry} $\spadesuit$}
        \label{fig:auc_description_insight_ellipsometry_population}
    \end{subfigure}
    \begin{subfigure}[b]{0.32\linewidth}
        \includegraphics[width=\linewidth, trim=0mm 0mm 0mm 0mm,clip]{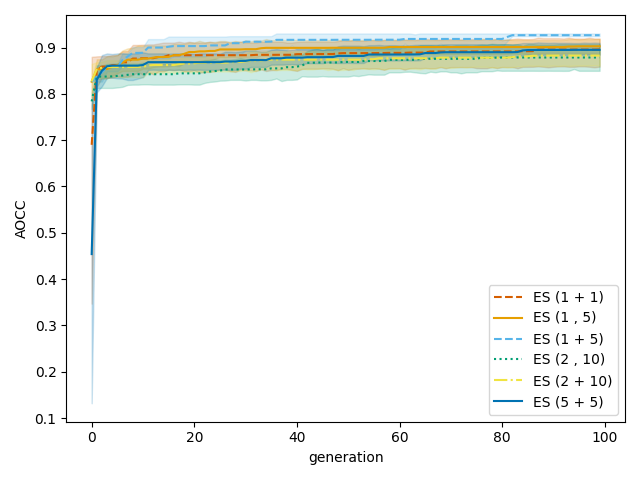}
        \caption{\textit{photovoltaic} $\spadesuit$}
        \label{fig:auc_description_insight_photovoltaic_population}
    \end{subfigure}
    \caption{Impact of different ES strategy choices. The preference for ES strategies is different for each problem and most of the time the difference is not significant.}
    \label{fig:ES_Exploration}
\end{figure*}

\input{input/convergency_curves}

However, in the photovoltaic problem, prompts without domain-specific insights outperformed the augmented variant, as shown in Fig.~\ref{fig:auc_description_insight_photovoltaic}. This counterintuitive result may be due to the the noisy fitness landscape of the photovoltaic task, where an overly detailed large prompt may prematurely limit exploration. Considering that all prompt settings in Fig.~\ref{fig:auc_description_insight_photovoltaic} start with a high AOCC, we let each LLaMEA run starting from the same solution to avoid the impact of initial gaps. However, the new and more reliable experimental results in Fig.~\ref{fig:auc_description_insight_photovoltaic_start_with_same_solution} still show that a large prompt may pose a limitation.

\subsection{Evolutionary Strategy Exploration}

The impact of the choice of evolutionary strategy on the performance of the algorithm varies according to the problem. In the Bragg mirror, as well as the ellipsometry problem, the (5+5) strategy (5 parents, 5 children) finally achieves the highest AOCC, as shown in Figs.~\ref{fig:auc_description_insight_bragg_population} and~\ref{fig:auc_description_insight_ellipsometry_population}, taking full advantage of the population diversity to exploit periodic solutions while retaining elite individuals for local improvement. In contrast, the photovoltaic problem favored the (1+5) strategy, as shown in Fig.~\ref{fig:auc_description_insight_photovoltaic_population}. Furthermore, while there are differences between the performance of these strategies, they are not significant.

\subsection{Algorithms Benchmarking}

\input{input/boxplot}
Fig.~\ref{fig:convergency_curves} shows the best 3 optimal algorithms found by LLaMEA for each of the 6 instances of 3 problems, compared to the commonly used algorithms. From Figs.~\ref{fig:convergency_minibragg} and~\ref{fig:convergency_bragg}, we find that the algorithms found by LLaMEA demonstrate a significantly more rapid convergence trend. However, the optimal algorithms found by LLaMEA for ellipsometry are all good at finding a local optimum at the initial stage, as shown in Fig.~\ref{fig:convergency_ellipsometry}. For the three instances of the photovoltaic problem, algorithms discovered by LLaMEA are not the best, but still show faster convergence.

Fig.~\ref{fig:boxplot} shows the distribution of the fitness value of the optimal solutions found by different algorithms for different problem instances after 15 runs. We find that the optimal algorithms found by LLaMEA perform very well and are basically comparable to the best algorithms. Especially for Bragg mirror problem instances, the found algorithms not only reliably find the optimal solutions but also the distribution is concentrated, which means that the performances are unfathomably stable. For photovoltaic problem instances, the algorithms found are relatively less stable but still reliable.

\section{Conclusions}
\label{sec:Conclusions}
This study demonstrates the potential of making real-world problem-related descriptions as well as algorithmic insights available to LLM and combining them with different evolutionary strategies for automatic algorithm discovery in photonic structure optimization. By embedding domain-specific knowledge into the structured task prompt, the algorithms generated by LLaMEA achieve advanced performance on real-world problems in comparison to other widely used successful algorithms in the area of photonic structure optimization. Key findings include:
\begin{itemize}
    \item Embedding domain-specific photonics knowledge into the LLM prompt, combined with evolutionary search strategies, enables automated discovery of optimization algorithms that achieve high performance in photonic structure design.
    \item In complex photonic optimization tasks, LLM generates algorithms that can match or even surpass state-of-the-art methods (e.g., DE and CMA-ES) with performance comparable to human-designed heuristics.
    \item Systematic tests using a variety of evolutionary strategies (e.g., (1 + 1), (5 + 5) and (2 + 10)) show that while different strategies affect the performance of a given problem (e.g., the (5 + 5) strategy performs well in the Bragg reflector and ellipsometry tasks while the (1+5) strategy performs moderately well in the photovoltaic task), the overall difference in performance is not significant.
    \item The algorithms found by LLM consistently find near-optimal solutions for each photonic design benchmark; in particular, they achieve very consistent results for Bragg mirror designs (reliably converging to the global optimum with small performance differences) and provide reliable results for challenging PV cases (with slightly larger differences, but still strong performance).
\end{itemize}

In summary, these findings confirm that LLM-driven automatic algorithm design can effectively automate the design of optimization algorithms for realistic photonic problems, paving the way for the fast and cost-effective design of photonic structures without the need for significant human intervention.

\FloatBarrier

\bibliographystyle{ACM-Reference-Format}
\bibliography{sample-base}

\newpage
\appendix

\section{Problem descriptions and algorithmic insights}

\begin{framed}{\textbf{Problem description}}
\begin{itemize}
    \item \textbf{The Bragg mirror optimization} aims to maximize reflectivity at a wavelength of 600 nm using a multilayer structure with alternating refractive indices (1.4 and 1.8). The structure's thicknesses are varied to find the configuration with the highest reflectivity. The problem involves two cases: one with 10 layers (minibragg) and another with 20 layers (bragg), with the latter representing a more complex inverse design problem. The known optimal solution is a periodic Bragg mirror, which achieves the best reflectivity by leveraging constructive interference. This case exemplifies challenges such as multiple local minima in the optimization landscape. 
    \item \textbf{The ellipsometry problem} involves retrieving the material and thickness of a reference layer by matching its reflectance properties using a known spectral response. The optimization minimizes the difference between the calculated and measured ellipsometric parameters for wavelengths between 400 and 800 nm and a fixed incidence angle of 40°. The parameters to be optimized include the thickness (30 to 250 nm) and refractive index (1.1 to 3) of the test layer. This relatively straightforward problem models a practical scenario where photonics researchers fine-tune a small number of parameters to achieve a desired spectral fit. 
    \item \textbf{The photovoltaics problem} optimizes the design of an antireflective multilayer coating to maximize the absorption in the active silicon layer of a solar cell. The goal is to achieve maximum short-circuit current in the 375 to 750 nm wavelength range. The structure consists of alternating materials with permittivities of 2 and 3, built upon a 30,000 nm thick silicon substrate. Three subcases with increasing complexity are explored, involving 10 layers (photovoltaics), 20 layers (bigphotovoltaics), and 32 layers (hugephotovoltaics). The optimization challenges include balancing high absorption with a low reflectance while addressing the inherent noise and irregularities in the solar spectrum. 
\end{itemize}
\end{framed}
\begin{framed}{\textbf{Algorithmic insight}}
\begin{itemize}
    \item \textbf{Bragg mirror}: For this problem, the optimization landscape contains multiple local minima due to the wave nature of the problem. And periodic solutions are known to provide near-optimal results, suggesting the importance of leveraging constructive interference principles. Here are some suggestions for designing algorithms: 1. Use global optimization algorithms like Differential Evolution (DE) or Genetic Algorithms (GA) to explore the parameter space broadly. 2. Symmetric initialization strategies (e.g., Quasi-Oppositional DE) can improve exploration by evenly sampling the search space. 3. Algorithms should preserve modular characteristics in solutions, as multilayer designs often benefit from distinct functional blocks. 4. Combine global methods with local optimization (e.g., BFGS) to fine-tune solutions near promising regions. 5. Encourage periodicity in solutions via tailored cost functions or constraints. 
    \item \textbf{Ellipsometry}: This problem has small parameter space with fewer variables (thickness and refractive index), and the cost function is smooth and relatively free of noise, making it amenable to local optimization methods. Here are suggestions for designing algorithms: 1. Use local optimization algorithms like BFGS or Nelder-Mead, as they perform well in low-dimensional, smooth landscapes. 2. Uniform sampling across the parameter space ensures sufficient coverage for initial guesses. 3. Utilize fast convergence algorithms that can quickly exploit the smooth cost function landscape. 4. Iteratively adjust bounds and constraints to improve parameter estimates once initial solutions are obtained. 
    \item \textbf{Photovoltaics}: This problem is a challenging high-dimensional optimization problem with noisy cost functions due to the realistic solar spectrum, and it requires maximizing absorption while addressing trade-offs between reflectance and interference effects. Here are the suggestions for designing algorithms: 1. Combine global methods (e.g., DE, CMA-ES) for exploration with local optimization for refinement. 2. Use consistent benchmarking and convergence analysis to allocate computational resources effectively. 3. Encourage algorithms to detect and preserve modular structures (e.g., layers with specific roles like anti-reflective or coupling layers). 4. Gradually increase the number of layers during optimization to balance problem complexity and computational cost. 5. Integrate robustness metrics into the cost function to ensure the optimized design tolerates small perturbations in layer parameters. 
\end{itemize}
    
\end{framed}

\end{document}

%% file: input/landscape.tex
\begin{figure}[!tb]
    \centering
    \begin{subfigure}[b]{0.325\linewidth}
        \includegraphics[width=\linewidth]{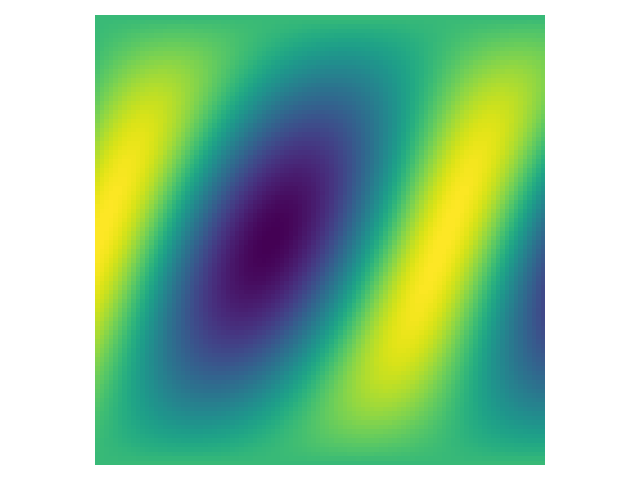}
        \caption{Bragg mirror}
        \label{fig:landscape_bragg}
    \end{subfigure}
    \begin{subfigure}[b]{0.325\linewidth}
        \includegraphics[width=\linewidth]{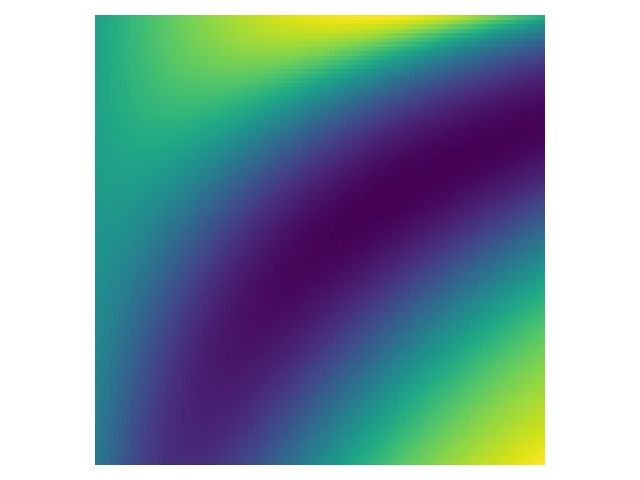}
        \caption{ellipsometry}
        \label{fig:landscape_ellipsometry}
    \end{subfigure}
    \begin{subfigure}[b]{0.325\linewidth}
        \includegraphics[width=\linewidth]{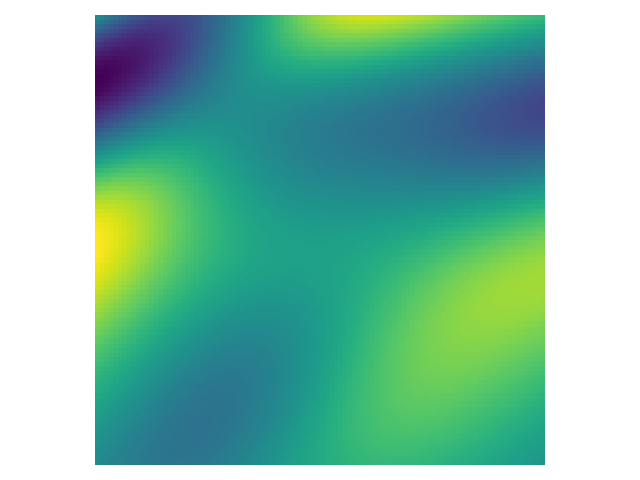}
        \caption{photovoltaics}
        \label{fig:landscape_photovoltaics}
    \end{subfigure}
    \caption{Landscape of photonic structure optimizing problems in 2D. The darker the color, the more fit the structure.}
    \label{fig:landscape}
\end{figure}

%% file: input/instances.tex
\begin{table*}[htb]
\centering
\caption{Parameter settings for different problem instances. Different columns of thickness and permittivity correspond to different materials. Algorithm discovery process for different problems are based on instances with $\spadesuit$ respectively.}
\label{tab:instances}
\resizebox{\textwidth}{!}{%
\begin{tabular}{@{}cccccclcccccc@{}}
\toprule
instances         & \multicolumn{2}{c}{\textit{mini-Bragg} $\spadesuit$}                & \multicolumn{2}{c}{\textit{Bragg}}                     & \multicolumn{2}{c}{\textit{ellipsometry} $\spadesuit$} & \multicolumn{2}{c}{\textit{photovoltaic} $\spadesuit$}            & \multicolumn{2}{c}{\textit{big-photovoltaic}}        & \multicolumn{2}{c}{\textit{huge-photovoltaic}}       \\ \midrule
layers            & \multicolumn{2}{c}{10}                        & \multicolumn{2}{c}{20}                        & \multicolumn{2}{c}{1}            & \multicolumn{2}{c}{10}                      & \multicolumn{2}{c}{20}                      & \multicolumn{2}{c}{32}                      \\
materials         & \multicolumn{2}{c}{2}                         & \multicolumn{2}{c}{2}                         & \multicolumn{2}{c}{1}            & \multicolumn{2}{c}{2}                       & \multicolumn{2}{c}{2}                       & \multicolumn{2}{c}{2}                       \\
min thickness(nm) & 0                     & 0                     & 0                     & 0                     & \multicolumn{2}{c}{50}           & 30                   & 30                   & 30                   & 30                   & 30                   & 30                   \\
max thickness(nm) & 218                   & 218                   & 218                   & 218                   & \multicolumn{2}{c}{150}          & 250                  & 250                  & 250                  & 250                  & 250                  & 250                  \\
min permittivity  & \multirow{2}{*}{1.96} & \multirow{2}{*}{3.24} & \multirow{2}{*}{1.96} & \multirow{2}{*}{3.24} & \multicolumn{2}{c}{1.1}          & \multirow{2}{*}{2.0} & \multirow{2}{*}{3.0} & \multirow{2}{*}{2.0} & \multirow{2}{*}{3.0} & \multirow{2}{*}{2.0} & \multirow{2}{*}{3.0} \\
max permittivity  &                       &                       &                       &                       & \multicolumn{2}{c}{3.0}          &                      &                      &                      &                      &                      &                      \\
evaluation budget & \multicolumn{2}{c}{10000}                     & \multicolumn{2}{c}{20000}                     & \multicolumn{2}{c}{1000}         & \multicolumn{2}{c}{5000}                    & \multicolumn{2}{c}{10000}                   & \multicolumn{2}{c}{16000}                   \\ \bottomrule
\end{tabular}%
}
\end{table*}

%% file: input/convergency_curves.tex
\begin{figure*}[!tb]
    \centering
    \begin{subfigure}[b]{0.325\linewidth}
        \includegraphics[width=\linewidth]{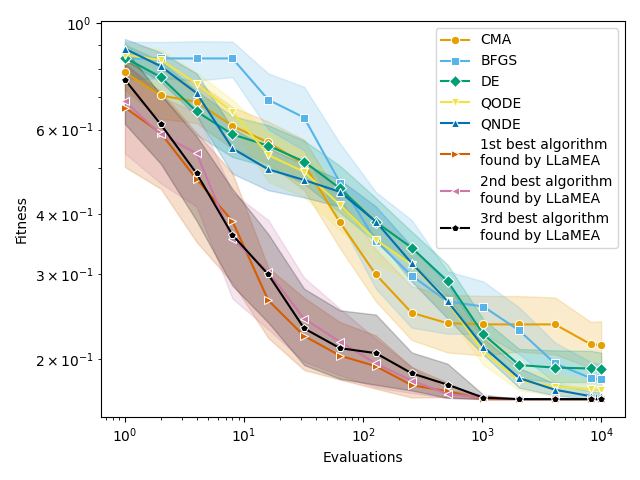}
        \caption{\textit{mini-Bragg} $\spadesuit$}
        \label{fig:convergency_minibragg}
    \end{subfigure}
    \begin{subfigure}[b]{0.325\linewidth}
        \includegraphics[width=\linewidth]{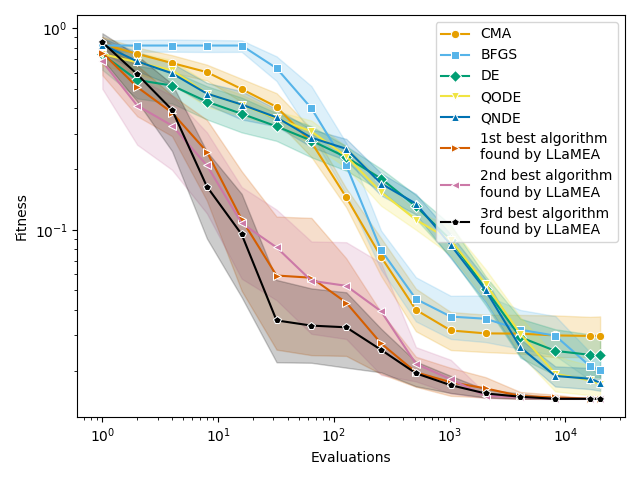}
        \caption{\textit{Bragg}}
        \label{fig:convergency_bragg}
    \end{subfigure}
    \begin{subfigure}[b]{0.325\linewidth}
        \includegraphics[width=\linewidth]{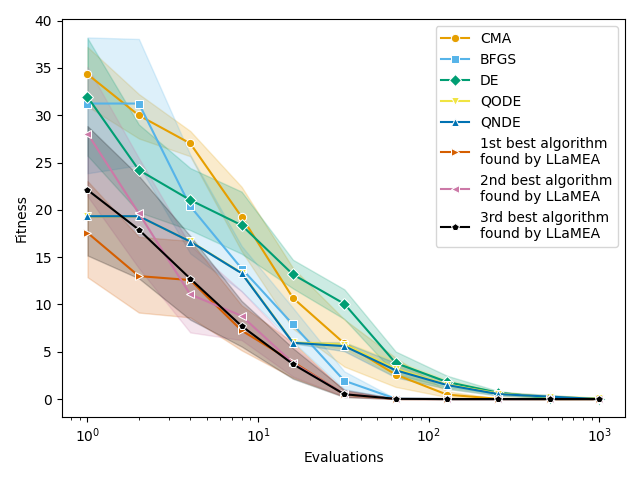}
        \caption{\textit{ellipsometry} $\spadesuit$}
        \label{fig:convergency_ellipsometry}
    \end{subfigure}
    \begin{subfigure}[b]{0.325\linewidth}
        \includegraphics[width=\linewidth]{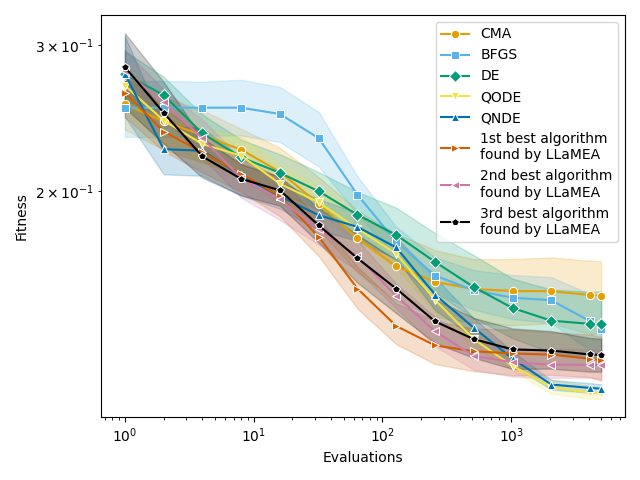}
        \caption{\textit{photovoltaic} $\spadesuit$}
        \label{fig:convergency_photovoltaics}
    \end{subfigure}
    \begin{subfigure}[b]{0.325\linewidth}
        \includegraphics[width=\linewidth]{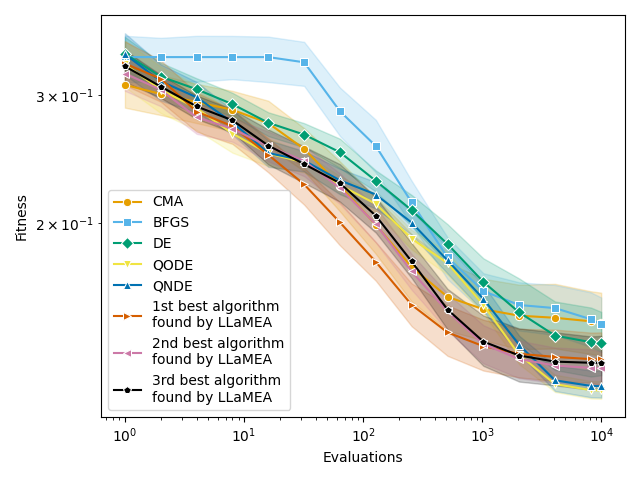}
        \caption{\textit{big-photovoltaic}}
        \label{fig:convergency_bigphotovoltaics}
    \end{subfigure}
    \begin{subfigure}[b]{0.325\linewidth}
        \includegraphics[width=\linewidth]{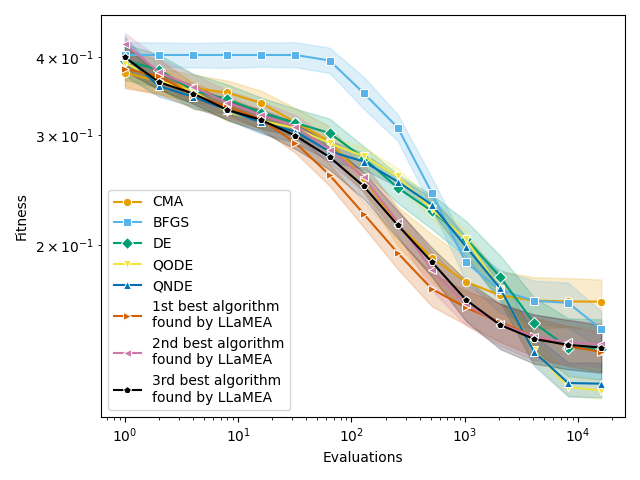}
        \caption{\textit{huge-photovoltaic}}
        \label{fig:convergency_hugephotovoltaics}
    \end{subfigure}
    \caption{Convergency curves of best algorithms found by LLaMEA and baselines with different problem instances, averaged over 15 runs. $y$-axis represents fitness, the smaller, the better. $x$-axis represents evaluations of problem instances. Each subfigure represents a problem instance.
    }
    \label{fig:convergency_curves}
\end{figure*}

%% file: input/boxplot.tex
\begin{figure*}[!ht]
    \centering
    \begin{subfigure}[b]{0.325\linewidth}
        \includegraphics[width=\linewidth]{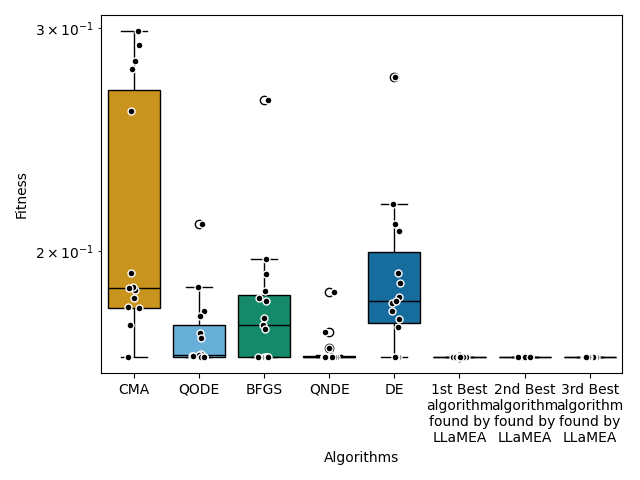}
        \caption{\textit{mini-Bragg} $\spadesuit$}
        \label{fig:boxplot_minibragg}
    \end{subfigure}
    \begin{subfigure}[b]{0.325\linewidth}
        \includegraphics[width=\linewidth]{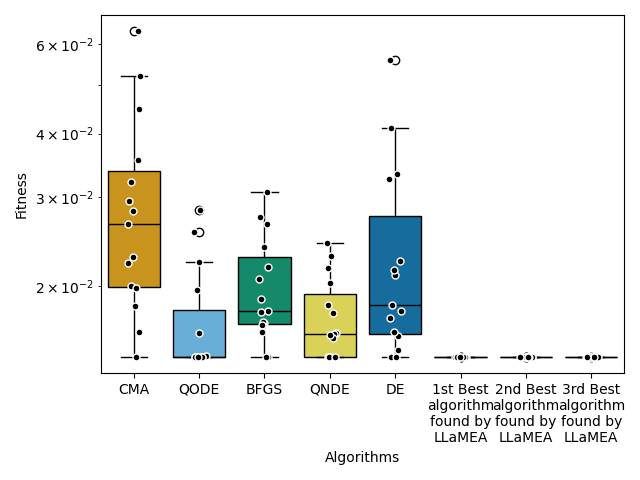}
        \caption{\textit{Bragg}}
        \label{fig:boxplot_bragg}
    \end{subfigure}
    \begin{subfigure}[b]{0.325\linewidth}
        \includegraphics[width=\linewidth]{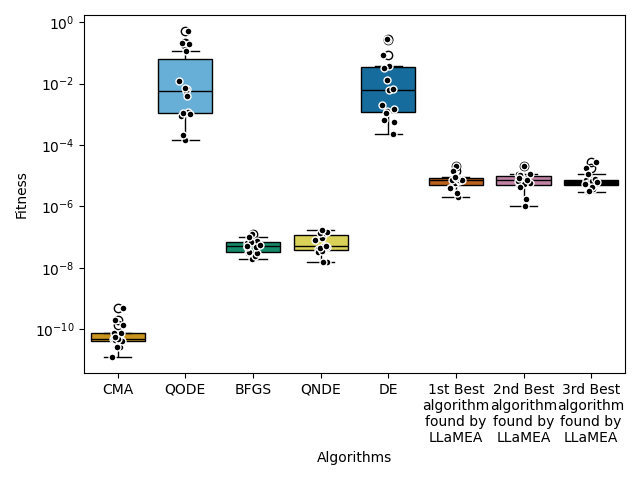}
        \caption{\textit{ellipsometry} $\spadesuit$}
        \label{fig:boxplot_ellipsometry}
    \end{subfigure}
    \begin{subfigure}[b]{0.325\linewidth}
        \includegraphics[width=\linewidth]{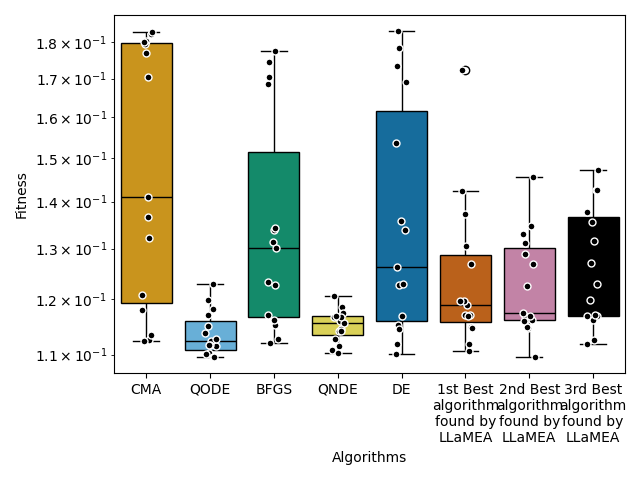}
        \caption{\textit{photovoltaic} $\spadesuit$}
        \label{fig:boxplot_photovoltaics}
    \end{subfigure}
    \begin{subfigure}[b]{0.325\linewidth}
        \includegraphics[width=\linewidth]{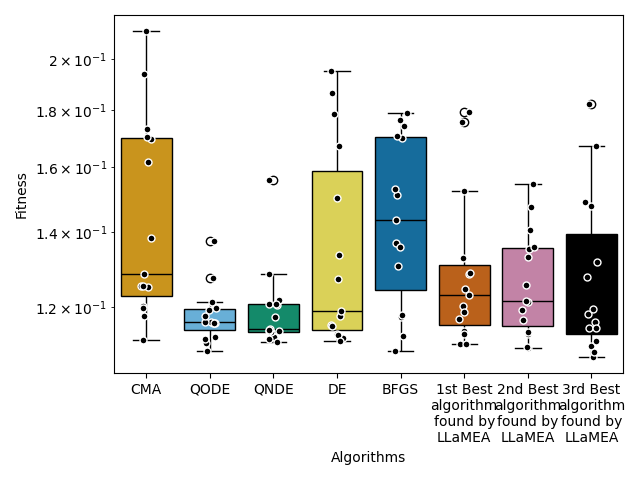}
        \caption{\textit{big-photovoltaic}}
        \label{fig:boxplot_bigphotovoltaics}
    \end{subfigure}
    \begin{subfigure}[b]{0.325\linewidth}
        \includegraphics[width=\linewidth]{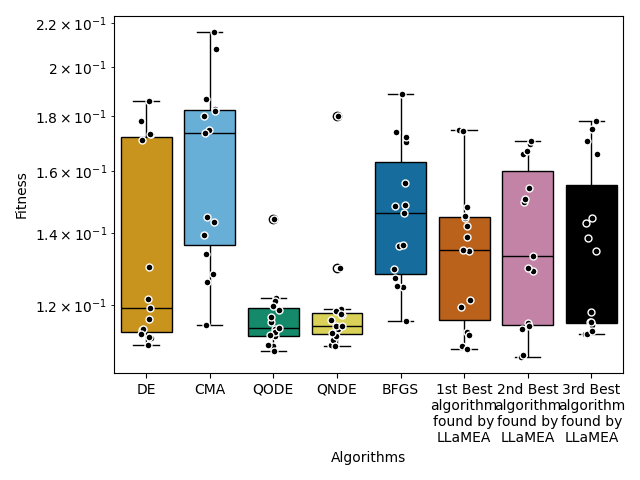}
        \caption{\textit{huge-photovoltaic}}
        \label{fig:boxplot_hugephotovoltaics}
    \end{subfigure}
    \caption{Distribution of best structure found by different algorithms. $y$-axis represent fitness of photonic structure. $x$-axis represent different algorithms. Each subfigure represents a problem instance.}
    \label{fig:boxplot}
\end{figure*}